\documentclass{WileyMSP-template}
\usepackage{hyperref}
\usepackage{svg}
\usepackage{orcidlink}
\usepackage{multirow}
\usepackage{booktabs} 
\usepackage{amsmath}
\usepackage{subcaption}
\usepackage{array} 

\usepackage{makecell}
\begin{document}

\pagestyle{fancy}

\title{A Retrieval-Augmented Knowledge Mining Method with Deep Thinking LLMs for Biomedical Research and Clinical Support}

\maketitle


\author{Yichun Feng}\orcidlink{0009-0007-4511-4713}
\author{Jiawei Wang}\orcidlink{0009-0005-0575-2695}
\author{Ruikun He*}
\author{Lu Zhou*}\orcidlink{0009-0003-6538-0649}
\author{Yixue Li*}\orcidlink{0000-0002-1198-7176}





\begin{affiliations}
Y. Feng\\
School of Advanced Interdisciplinary Sciences  \\
University of Chinese Academy of Sciences  \\
Beijing 100049, China  \\
\vspace{1em} 

Y. Feng, Y. Li  \\
Key Laboratory of Systems Health Science of Zhejiang Province\\  
School of Life Science\\ 
Hangzhou Institute for Advanced Study\\ 
University of Chinese Academy of Sciences  \\
Hangzhou 310024, China  
\vspace{1em} 

Y. Feng, L. Zhou, Y. Li \\ 
Guangzhou National Laboratory  \\
No. 9 XingDaoHuanBei Road, Guangzhou International Bio Island\\
Guangzhou 510005, China  \\
E-mail: zhou\_lu@gzlab.ac.cn  \\
E-mail: li\_yixue@gzlab.ac.cn 
\vspace{1em} 

J. Wang  \\
Department of EEIS  \\
University of Science and Technology of China  \\
Hefei 230026, China  
\vspace{1em} 

R. He \\
BYHEALTH Institute of Nutrition \& Health\\
Guangzhou 510663, China\\
E-mail: herk@by-health.com

\vspace{1em} 
Y. Li \\ 
GZMU-GIBH Joint School of Life Sciences\\ The Guangdong-Hong Kong-Macau Joint Laboratory for Cell Fate Regulation and Diseases\\
Guangzhou Medical University\\ 
Guangzhou, 511436, China

\vspace{1em} 
Y. Li \\ 
School of Life Sciences and Biotechnology\\
Shanghai Jiao Tong University\\
Shanghai 200240, China

\vspace{1em} 
Y. Li \\ 
Shanghai Institute of Nutrition and Health\\
Chinese Academy of Sciences\\
Shanghai 200030, China

\end{affiliations}


\keywords{Large Language Model, Knowledge Graph, Knowledge Mining, Retrieval-Augmented Generation, Deep Thinking}

\vspace{1em} 
\begin{abstract}

Knowledge graphs and large language models (LLMs) are key tools for biomedical knowledge integration and reasoning, facilitating structured organization of scientific articles and discovery of complex semantic relationships. However, current methods face challenges: knowledge graph construction is limited by complex terminology, data heterogeneity, and rapid knowledge evolution, while LLMs show limitations in retrieval and reasoning, making it difficult to uncover cross-document associations and reasoning pathways.
To address these issues, we propose a pipeline that uses LLMs to construct a biomedical knowledge graph (BioStrataKG) from large-scale articles and builds a cross-document question-answering dataset (BioCDQA) to evaluate latent knowledge retrieval and multi-hop reasoning. We then introduce Integrated and Progressive Retrieval-Augmented Reasoning (IP-RAR) to enhance retrieval accuracy and knowledge reasoning. IP-RAR maximizes information recall through Integrated Reasoning-based Retrieval and refines knowledge via Progressive Reasoning-based Generation, using self-reflection to achieve deep thinking and precise contextual understanding. 
Experiments show that IP-RAR improves document retrieval F1 score by 20\% and answer generation accuracy by 25\% over existing methods. This framework helps doctors efficiently integrate treatment evidence for personalized medication plans and enables researchers to analyze advancements and research gaps, accelerating scientific discovery and decision-making.

\end{abstract}


\section{Introduction}
The advancement of large language models (LLMs) has significantly accelerated progress in natural language processing (NLP), particularly in complex tasks like question answering (QA), with notable breakthroughs in specialized fields such as biomedical science \cite{luo2024taiyi}. However, despite these advancements, the application of LLMs for knowledge mining in specialized domains like biomedical science remains limited, where rigorous precision and robust evidence validation are essential for extracting meaningful insights \cite{liu2021towards}. Biomedical articles is vast, containing massive amounts of information \cite{comeau2013bioC}, much of which remains underutilized. This inefficiency hinders scientific discovery and limits our ability to address complex biomedical challenges that require deep connections and advanced reasoning \cite{doan2014natural}. The wealth of knowledge within biomedical research, if effectively harnessed, could drive breakthroughs in critical areas such as clinical medicine, pharmacology, and molecular biology \cite{cai2023integrating}.
\vspace{1em}

In biomedical knowledge mining, the construction of knowledge graphs is essential for enhancing information integration and reasoning capabilities. However, this process presents several challenges. First, the biomedical domain has a highly complex terminology system, with numerous synonyms, polysemous terms, and hierarchical classifications, making the precise extraction of entities and relationships particularly difficult \cite{chang2020benchmark}. Second, biomedical knowledge is fragmented and heterogeneous, spanning research papers, clinical reports, and databases, posing a fundamental challenge in unifying and linking these diverse data sources effectively \cite{li2020real}. Additionally, the rapid evolution of biomedical research constantly introduces new findings, requiring knowledge graphs to be dynamically expandable to maintain long-term relevance \cite{zheng2021pharmkg}. Despite these challenges, high-quality knowledge graphs can explicitly capture the intricate relationships among genes, proteins, diseases, and drugs, providing structured support for cross-document reasoning and deep knowledge extraction \cite{wu2023medical}.
\vspace{1em}

Building a high-quality biomedical QA dataset that effectively leverages knowledge graphs is crucial for evaluating and enhancing models’ abilities in cross-document reasoning and knowledge integration. While existing datasets such as MASH-QA \cite{zhu2020question}, which addresses multi-span questions across long documents, BioASQ \cite{nentidis2023overview}, which focuses on biomedical semantic indexing and QA, MEDHOP \cite{welbl2018constructing}, which targets multi-hop reasoning across multiple paragraphs, MedicationQA \cite{abacha2019bridging}, which enhances understanding of medication-related queries, MedMCQA \cite{pal2022medmcqa}, which provides large-scale multiple-choice QA for medical examinations, and PcQA \cite{Feng2024}, which facilitates structured knowledge graph question answering, offer valuable benchmarks, they often fail to capture the deep, interconnected knowledge hidden within multiple sources. 
\vspace{1em}

Retrieval-Augmented Generation (RAG) has emerged as a promising approach for biomedical knowledge mining by integrating external knowledge into the response generation process \cite{lewis2020retrieval}. However, its effectiveness relies on refining the retrieval mechanism to focus on high-quality, contextually relevant content, ensuring more accurate and reliable knowledge extraction  \cite{jegal2023learning}. To address this, SELF-RAG \cite{asai2023self} dynamically adjusts the retrieval process by evaluating the quality of the retrieved content, thereby enhancing the accuracy of long-form answers in complex domains like biomedical research. Other innovative approaches include GraphRAG \cite{edge2024local}, which structures retrieved passages into graphs to capture relationships between them, improving coherence and relevance. CRAG \cite{yan2024corrective} further refines the process by iteratively filtering out irrelevant passages, ensuring only the most pertinent information is used. Additionally, RAPTOR \cite{sarthi2024raptor} organizes retrieval results into a tree structure, recursively summarizing data to facilitate more effective reasoning.
\vspace{1em}

In addition to extracting high-quality information from external knowledge bases using RAG, LLMs have recently made breakthrough progress in their inherent reasoning capabilities. For instance, OpenAI's GPT‑o1 leverages reinforcement learning and a chain-of-thought mechanism to demonstrate reasoning on complex tasks in mathematics, programming, and science—achieving performance comparable to that of PhD-level experts \cite{jaech2024openai}. Meanwhile, DeepSeek has released its reasoning-oriented model, DeepSeek‑R1 \cite{guo2025deepseek}, which performs on par with GPT‑o1 in logical reasoning, mathematical computation, and code generation, yet its training cost is only a small fraction of that of GPT‑o1. These breakthroughs not only expand the functionalities of large models but also provide stronger intrinsic support for complex domains such as biomedical knowledge mining, helping to yield more accurate answers in cross-document and multi-layer relational reasoning tasks.
\vspace{1em}

\begin{figure}[bt!]
\centering
\includegraphics[width=1\linewidth]{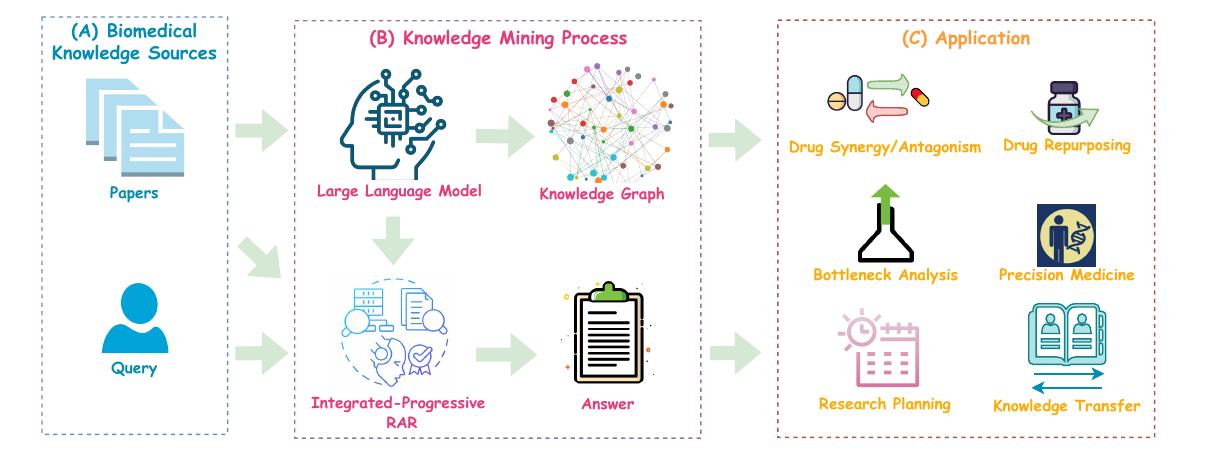}
\caption{\label{fig:application} 
Overview of the proposed framework for biomedical knowledge mining. (A) Biomedical knowledge sources, such as research papers and user queries, are processed through (B) 
A knowledge mining pipeline that leverages a LLM alongside the IP-RAR approach, designed to generate knowledge graph and precise answers. (C) The outputs enable diverse applications, including drug synergy/antagonism, drug repurposing, precision medicine, bottleneck analysis, research planning, and knowledge transfer.}
\end{figure}

In this paper, we present a comprehensive framework for biomedical knowledge mining, as illustrated in Figure \ref{fig:application}. This framework processes diverse biomedical knowledge sources, such as research papers and user queries, through a systematic pipeline that involves constructing knowledge graphs and generating precise responses to user queries. Within this pipeline, LLMs collaborate with retrieval-augmented reasoning techniques to extract, organize, and integrate domain-specific knowledge. First, an entity-level knowledge graph and a document-level knowledge graph are constructed from large-scale biomedical research papers and merged into the Biomedical Stratified Knowledge Graph (BioStrataKG). The entity-level graph captures relationships between biomedical entities such as genes, proteins, diseases, and drugs, while the document-level graph represents connections between research papers based on shared methods, datasets, and research directions, as well as citation and reference relationships. Together, these knowledge graphs uncover complex biomedical relationships and reveal latent patterns, thereby supporting advanced knowledge discovery. 
Since existing datasets do not meet the requirements for cross-document reasoning and knowledge mining, we develop a new biomedical cross-document question answering dataset, called BioCDQA, based on the BioStrataKG to support these tasks.
Second, we propose the Integrated and Progressive Retrieval-Augmented Reasoning (IP-RAR) framework, which synergizes Integrated Reasoning-based Retrieval with Progressive Reasoning-based Generation to seamlessly integrate relevant knowledge, enabling more precise and contextually relevant reasoning. The framework first maximizes the recall of pertinent information from large-scale biomedical articles through Integrated Reasoning-based Retrieval, ensuring comprehensive coverage of relevant data. Subsequently, the progressive reasoning-based generation mechanism refines and enhances the extracted knowledge while leveraging a self-reflection mechanism to continuously optimize the accuracy and contextual relevance of the answers. Ultimately, LLMs with deep thinking capability further refine the reasoning process to derive a high-quality answer.
The IP-RAR framework enables precise reasoning and knowledge integration, providing effective technical support for a wide range of downstream biomedical research applications, including drug synergy/antagonism analysis, drug repurposing, precision medicine, and knowledge transfer. This framework helps doctors quickly identify and integrate relevant treatment evidence from vast biomedical articles, enabling more precise personalized medication plans. It also allows researchers to systematically analyze cutting-edge advancements and potential research gaps, accelerating research strategy formulation and decision-making. These advancements in knowledge mining pave the way for more efficient and accurate solutions in contemporary biomedical research.

\section{Results}
\subsection{Construction of the BioStrataKG}
This section introduces the construction method of the BioStrataKG, proposing a document-entity dual-layer representation fusion architecture based on LLMs, which effectively captures complex relationship networks among biomedical entities and establishes a cross-document knowledge association system based on this network. As shown in Figure \ref{fig:dataset_construction}, the construction process of BioStrataKG begins with large-scale biomedical articles, utilizing GPT-4o mini \cite{openai_gpt-4o-mini_2024} for fine-grained knowledge extraction, including entity-relationship triple extraction and the structured representation of semantic information such as research methods and fields in the article. Subsequently, cross-document association networks are established through entity co-occurrence analysis, achieving a hierarchical expansion of the knowledge graph from micro-level entity relationships to macro-level document associations. The entity types and relationship types within the knowledge graph are detailed in Figure \ref{fig:rag_KG}.

\subsection{Overview of the Biomedical Cross-Document Question Answering Dataset}

\begin{figure}[bt!]
    \centering
    \begin{subfigure}[b]{\textwidth}
        \centering
        \includegraphics[width=1\textwidth]{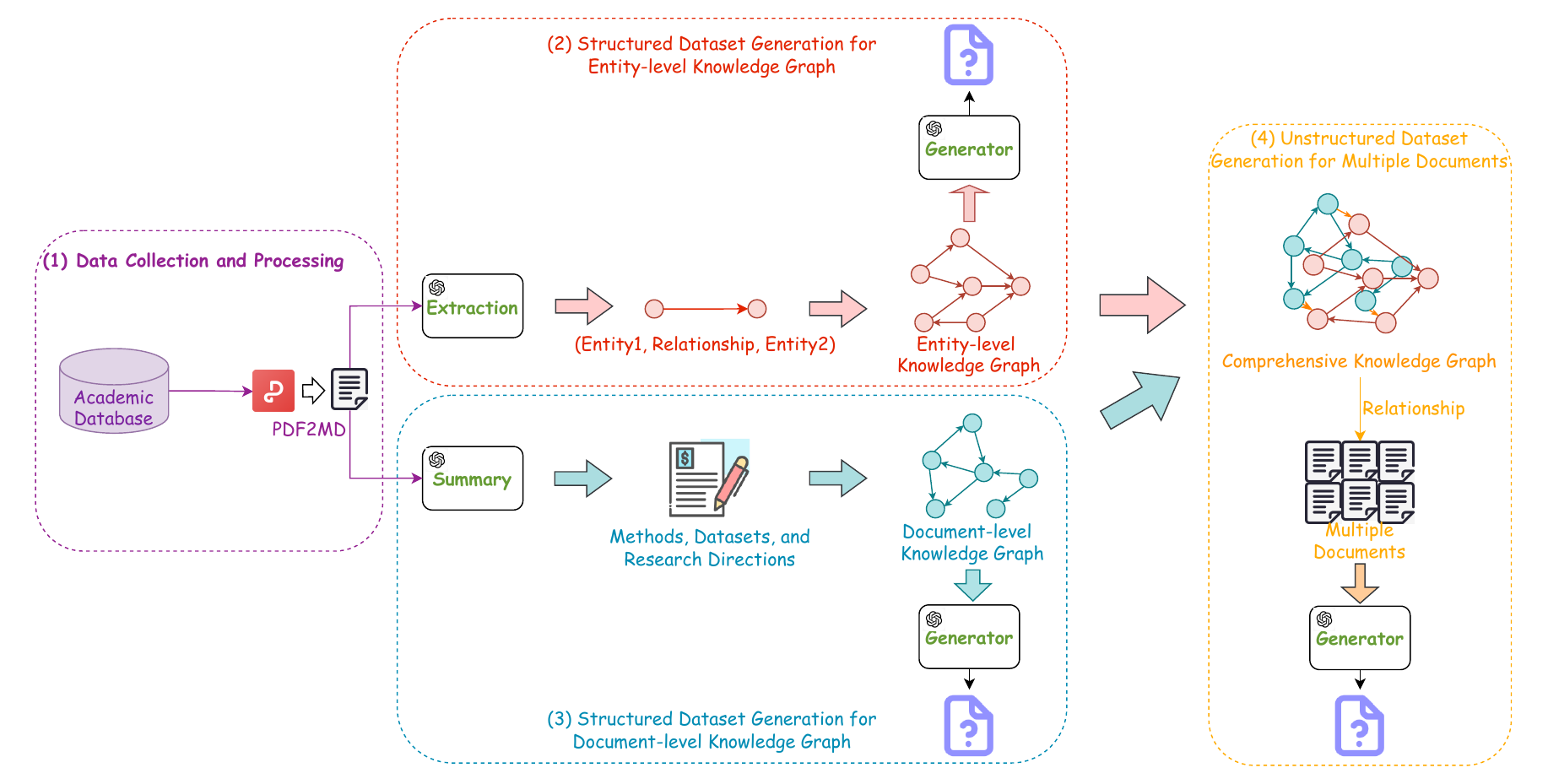} 
        \caption{} 
        \label{fig:dataset_construction}
    \end{subfigure}
    
    \begin{subfigure}[t]{0.49\textwidth}
        \centering
        \raisebox{1cm}{\includegraphics[width=\textwidth]{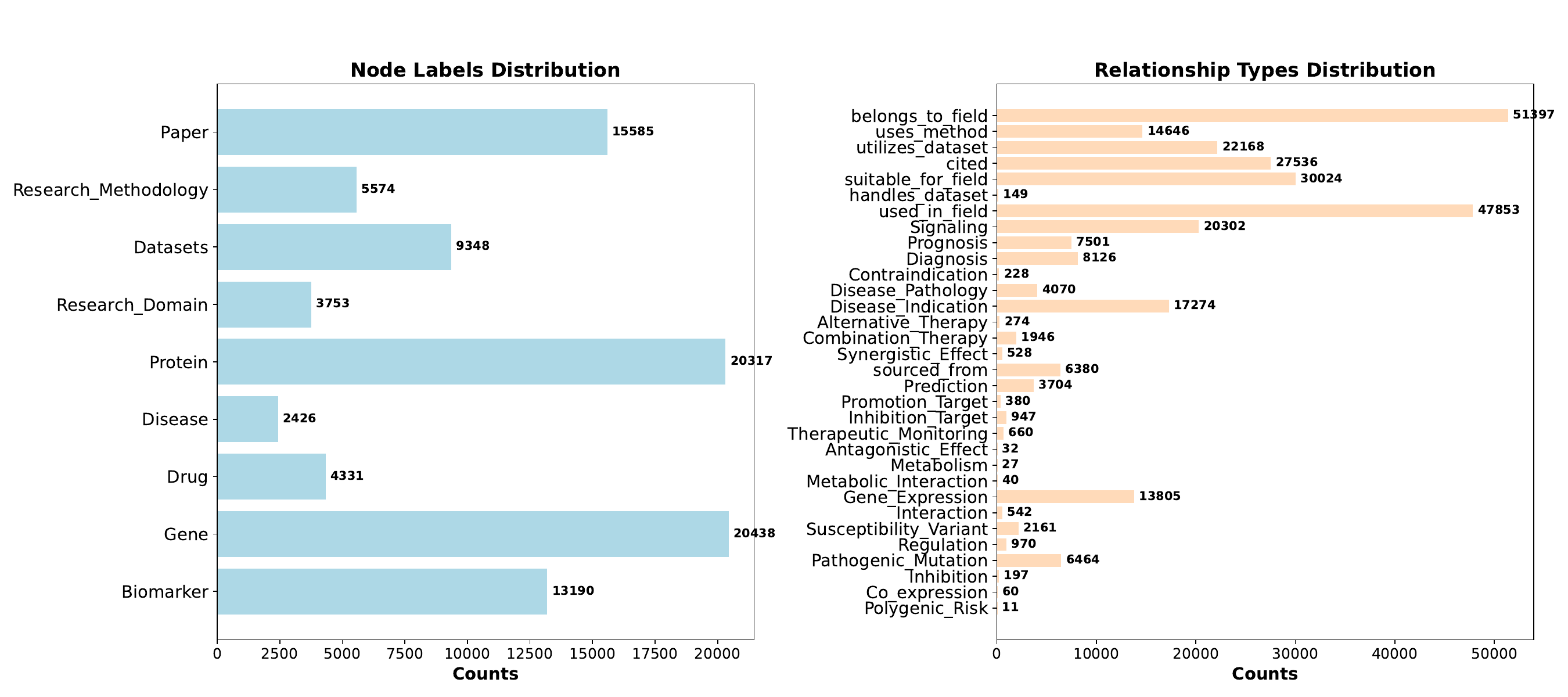}}
        \caption{}
        \label{fig:rag_KG}
    \end{subfigure}
    \hfill
    \begin{subfigure}[t]{0.49\textwidth}
        \centering
        \includegraphics[width=\textwidth]{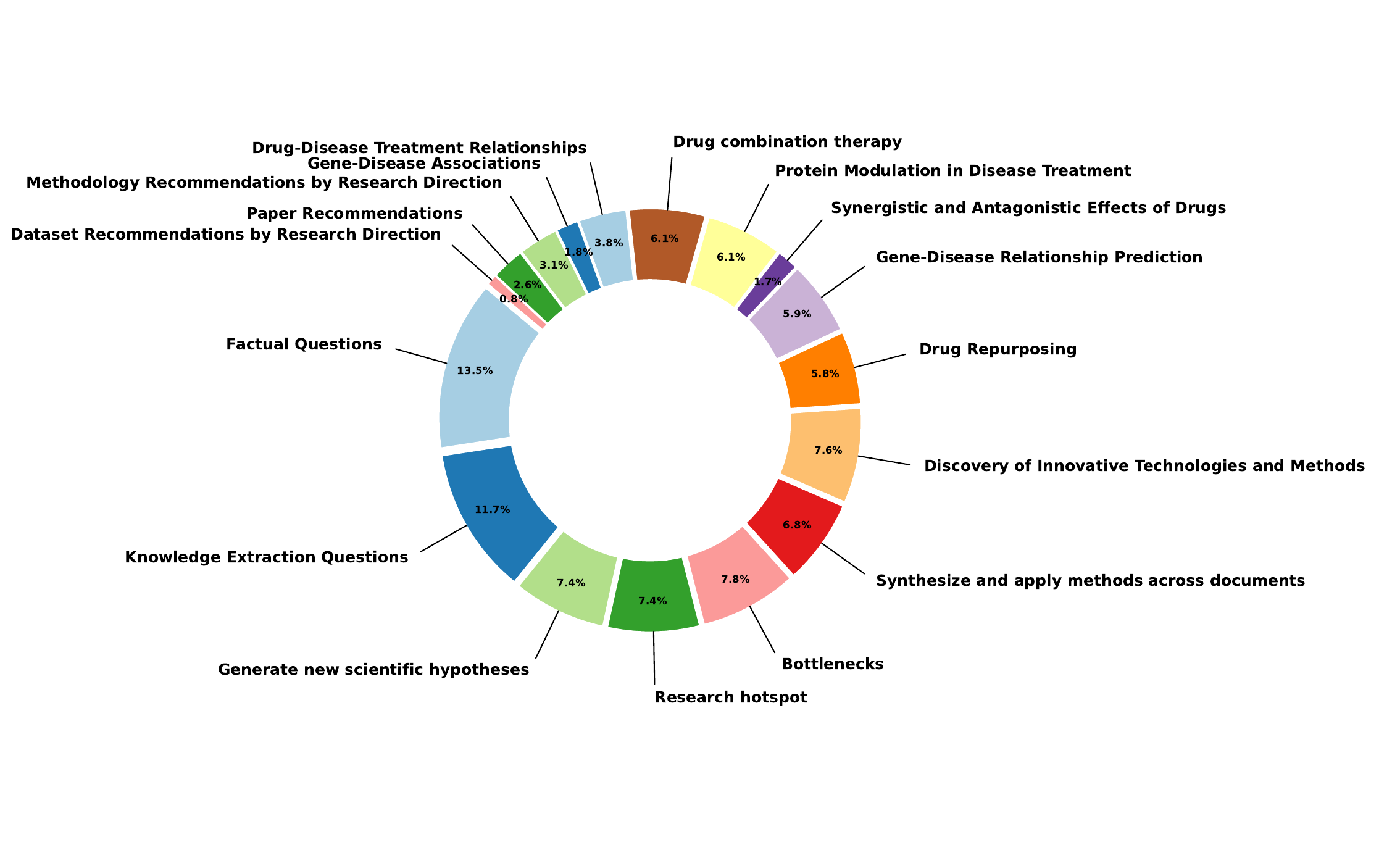}
        \caption{}
        \label{fig:rag_dataset}
    \end{subfigure}
    
    \caption{\label{fig:Statistical Analysis}Construction Pipeline and Statistical Analysis of the Dataset. (a) BioStrataKG and BioCDQA Construction Workflow Diagram.
        (1) Data Collection and Processing: The process begins by converting research papers from PDF to markdown (MD) format to facilitate content extraction.
(2) Structured Dataset Generation for Entity-level Knowledge Graph: An LLM is used to extract entities and relationships (Entity1, Relationship, Entity2), which are then standardized to construct an entity-level knowledge graph. This graph supports downstream tasks such as drug repurposing, drug interaction analysis for comorbid conditions, and gene-disease associations.
(3) Structured Dataset Generation for Document-level Knowledge Graph: Summarization is performed using an LLM to extract key aspects such as methods, datasets, and research directions. The resulting document-level knowledge graph facilitates tasks such as research strategic planning and research paper recommendations.
(4) Unstructured Dataset Generation for Multiple Documents: Integration of the entity-level and document-level knowledge graphs produces a comprehensive knowledge graph. This integrated graph enables connections across multiple documents and supports downstream tasks such as content-based factual questioning, research bottleneck analysis, knowledge transfer, trend analysis, and hotspot detection. (b) Statistics of node labels and relationship types in BioStrataKG. (c) Statistics of question categories in BioCDQA.} 
\end{figure}

 Based on BioStrataKG, we introduce a biomedical question-answering dataset—BioCDQA, designed to support cross-document reasoning and biomedical knowledge mining. The construction process of the dataset is illustrated in Figure \ref{fig:dataset_construction}. The dataset integrates data from both unstructured text and structured knowledge graphs. By extracting information from text, we obtain rich contextual data, while the knowledge graph provides precise relationships between entities such as genes, diseases, drugs, and proteins. These data sources collectively ensure the diversity and practicality of the dataset. This dataset consists of tuples containing the following elements: question, question type, answer, source papers for the answer, and source sentences for the answer. Each tuple includes a natural language question, whose answer is composed of one or more sentences extracted from the source papers. These answers may originate from a single paper or multiple papers and can consist of multiple sentences, either from a single span or dispersed across different sections of various source papers.
\vspace{1em} 

The dataset contains a total of 1,183 question-answer pairs, covering 68,428 papers and providing over 1.85 million document chunks available for retrieval. The dataset and the corresponding set of retrievable papers will soon be made publicly available.
The distribution of question types is shown in Figure \ref{fig:rag_dataset}. The definitions and characteristics of each type are elaborated in the subsequent sections.

\subsubsection{Question Types from Unstructured Data}
The question types generated from unstructured biomedical articles aim to comprehensively uncover both explicit and implicit knowledge within the text, supporting reasoning tasks ranging from basic fact retrieval to complex scientific discovery. This dataset encompasses the following question types:

\threesubsection{Factual Questions}
In the domain of biology, factual questions typically revolve around specific biological facts, data, or phenomena. Examples include inquiries about the sequence of a particular gene, the function of a specific protein, epidemiological data of a disease, or the detailed steps of a biological process. These questions can be answered through a thorough examination of scientific article, experimental data, or validated biological databases, ensuring responses are grounded in objective, verifiable information rather than subjective opinions.

\threesubsection{Knowledge Extraction Questions}
The primary objective of knowledge extraction questions is to identify and extract specific categories of information from a given text, typically presented in a list format. In the biomedical domain, such questions may involve extracting names of drugs, genes, diseases, or symptoms. For instance, when asked, "What are the drugs used to treat lung cancer?", the system must extract relevant drug names from the text and present them in a list, typically identifying three to five commonly used drugs. The focus of these questions is on precise extraction rather than interpretation or reasoning. The goal is to quickly distill key entity information from large volumes of unstructured text, enabling researchers to efficiently retrieve the necessary biomedical data and accelerate literature analysis.

\threesubsection{Knowledge Discovery Questions}
Knowledge discovery questions represent a key feature of our dataset, aiming to extract deeper insights and foster innovative scientific inquiry. These questions not only facilitate the generation of new hypotheses from the findings and conclusions presented in the articles but also assist in designing subsequent research plans. By analyzing existing articles, these questions help identify hidden themes and emerging research hotspots, providing valuable insights into current trends. Additionally, they empower researchers to analyze publication trends, assess keyword relevance, and evaluate citation metrics, uncovering critical areas of interest and guiding potential future directions in biomedical research. Moreover, knowledge discovery questions allow for the identification of bottlenecks within specific fields, based on comprehensive reviews, which highlights pressing challenges that need addressing. Through the synthesis of information across multiple documents, these questions support the transfer of methodologies and concepts to other research areas, fostering interdisciplinary innovation. Overall, knowledge discovery questions play a pivotal role in advancing the scientific dialogue by tracking the evolution of knowledge, revealing critical nodes in research development, and ultimately guiding future investigations in the biomedical domain.\\

\subsubsection{Question Types from Structured Data}
The question types based on BioStrataKG focus on the systematic reasoning and prediction of complex relationships among biomedical entities, supporting multi-level research needs ranging from molecular mechanism exploration to clinical treatment optimization. This dataset covers the following core task directions:

\threesubsection{Drug Repurposing} Triplet relationships between diseases, genes, and drugs are established to uncover new applications for existing drugs in treating previously unrelated diseases. By leveraging data on drug effects and identifying hidden connections, innovative treatment pathways are discovered.

\threesubsection{Gene-Disease Relationship Prediction} Known associations between genes and diseases are analyzed to predict new gene-disease relationships. Patterns and similarities within the data highlight potential genetic markers involved in disease development, aiding early diagnosis and targeted therapies.

\threesubsection{Synergistic and Antagonistic Effects of Drugs} Drug interactions are examined with a focus on synergistic (enhancing) or antagonistic (counteracting) effects. For patients with comorbid conditions, harmful interactions are identified, and safer alternatives are suggested, improving patient safety and optimizing treatment effectiveness.

\threesubsection{Protein Modulation in Disease Treatment} The inhibitory or promotive effects of proteins on disease management are analyzed, providing insights into protein-drug interactions and supporting the design of effective therapeutic strategies.

\threesubsection{Drug combination therapy} Drug combination therapies are analyzed to optimize dosages and interactions for treating complex conditions, ensuring maximum therapeutic benefits with minimal side effects.

\threesubsection{Drug-Disease Treatment Relationships} Relationships between specific drugs and the diseases they treat are explored to gain insights into effective treatment options and potential new therapeutic applications for existing drugs.

\threesubsection{Gene-Disease Associations} Comprehensive information on gene-disease associations reveals genetic predispositions, enabling the development of genetic tests for early detection and personalized treatments based on individual genetic profiles.

\threesubsection{Dataset Recommendations by Research Direction} Metadata and articles are analyzed to identify and recommend datasets that align with specific research goals, providing high-quality, curated resources tailored to researchers' needs.

\threesubsection{Methodology Recommendations by Research Direction} Suitable research methodologies are recommended based on the identified direction, equipping researchers with effective approaches ranging from traditional techniques to emerging methods.

\threesubsection{Paper Recommendations by Research Direction} Relevant papers are suggested based on research direction, including seminal works, recent publications, and highly cited studies, enabling researchers to build on existing knowledge and stay updated on advancements in their field.\\

\subsection{Overview of IP-RAR Framework}

\begin{figure}[t!]
\centering
\includegraphics[width=1\linewidth]{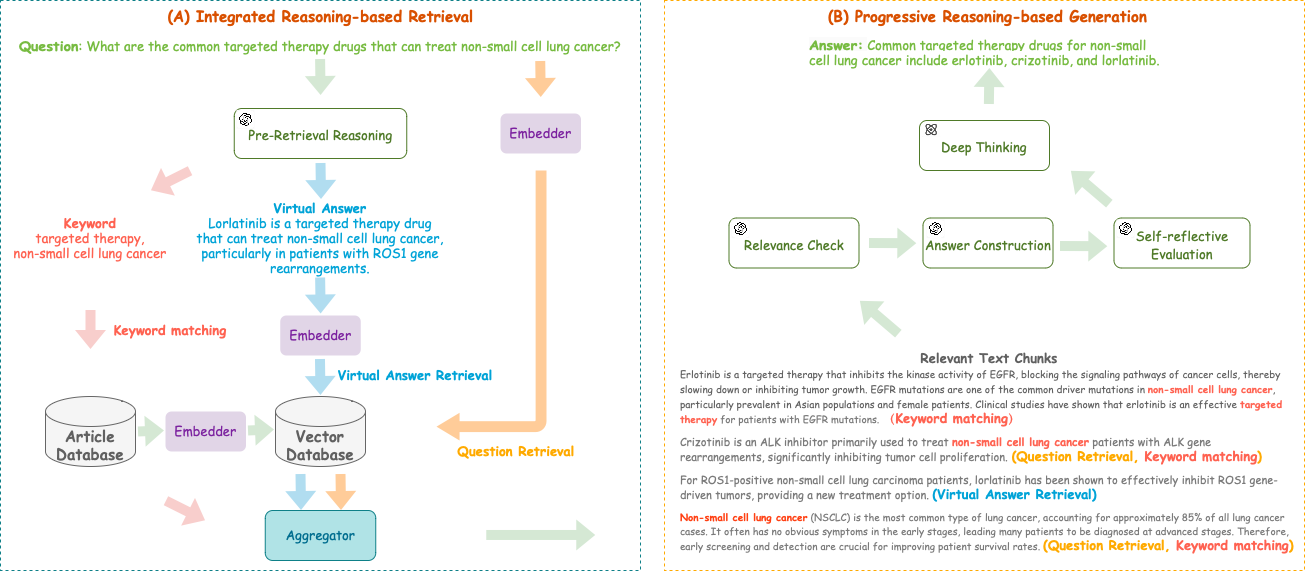}
\caption{\label{fig:frameworks}Framework of IP-RAR. (A) Integrated Reasoning-based Retrieval: Performs pre-retrieval reasoning, extracting keywords and generating a virtual answer. Then, a multi-level, multi-granularity retrieval strategy is used to retrieve relevant text chunks, which are ranked based on relevance.
(B) Progressive Reasoning-based Generation: Filters out irrelevant text chunks through explanations or self-reflection, then leverages DeepSeek-R1 for deep-thinking-based reasoning on the valid text chunks, generating a precise final response.}
\end{figure}

In this study, we propose the IP-RAR framework, specifically designed for deep-thinking-based knowledge mining and question answering in large-scale biomedical articles. The overall framework is illustrated in Figure~\ref{fig:frameworks}.
In the Integrated Reasoning-based Retrieval phase, our approach first performs pre-retrieval reasoning, leveraging DeepSeek-V3 \cite{liu2024deepseek} to analyze the semantic information of the input question, extract keywords, and generate an initial virtual answer. Then, using Contriever-MS MARCO text embedding model \cite{izacard2021unsupervised}, a multi-level and multi-granularity retrieval strategy is proposed to retrieve relevant text chunks from 68,428 research papers (approximately 1.85 million chunks) based on the input question, the extracted keywords, and the initial virtual answer. These text chunks are then ranked by an aggregator to ensure high relevance.
In the Progressive Reasoning-based Generation phase, we utilize DeepSeek-V3 to filter out irrelevant text chunks through explanation or self-reflection and selects the most informative ones. These selected text chunks are used to generate individual answers, which are then further validated to determine whether they can accurately address the original question. Finally, our framework integrates the question with the corresponding text chunks of valid answers and leverages the DeepSeek-R1 \cite{guo2025deepseek} to perform deep-thinking-based reasoning, producing a comprehensive and precise final response.
Details can be found in Section \ref{sec:IP-RAR}.

\subsection{Results of Various Approaches on the BioCDQA Dataset}
For the BioCDQA dataset, we design a set of evaluation metrics. Document retrieval performance is assessed using Mean Precision, Mean Recall, and Mean F-measure to evaluate retrieval accuracy and completeness. Answer accuracy evaluation relies on a GPT-4-based five-point scoring system, ensuring a comprehensive assessment of the semantic consistency and precision of generated answers. This approach is particularly suitable for summary-type question-answering tasks, where traditional metrics struggle to provide stable quality measurements. A detailed introduction to the evaluation metrics can be found in the Supporting Information section 1.1.
\vspace{1em} 

As shown in Table~\ref{tab:bioCDQA}, the IP-RAR system demonstrates superior performance in biomedical cross-document question answering. During the document retrieval stage, the system achieves an F1-score of 34.96\%, significantly outperforming baseline methods by a relative margin of 12-18 percentage points. Its balanced performance between precision and recall ensures high-quality input for subsequent answer generation. In contrast, existing approaches such as SELF-RAG, Contriever+Llama3, and CRAG exhibit substantially lower F1-scores (all below 22\%), indicating insufficient retrieval effectiveness that propagates errors to downstream tasks.
\vspace{1em} 

In the answer generation stage, IP-RAR leads with the highest GPT-4 evaluation score of 76.41\%, demonstrating the effectiveness of its generation strategies. 
\vspace{1em}

Overall, IP-RAR demonstrates significant advantages in both retrieval and answer generation, outperforming all existing systems comprehensively.

\begin{table*}[t]
    \centering
    \begin{tabular}{lcccccccccc}
    
    \toprule
    \multirow{2}{*}{\textbf{System}} & \multicolumn{3}{c}{\textbf{Document Retrieval}} & \textbf{Answer} \\
    \cmidrule(lr){2-4} \cmidrule(lr){5-5}
     & \textbf{Prec. (\%)} & \textbf{Rec. (\%)} & \textbf{F1 (\%)}  &  \textbf{GPT-4 Eval (\%)} \\
    \midrule
    Contriever+Llama 3 8B&24.53&19.40&21.67& 25.86\\
    CRAG&12.18&18.49&14.69&29.27\\
    SELF-RAG & 7.12& \textbf{31.28}&11.59 &41.16 \\
    \textbf{IP-RAR(ours)} & \textbf{47.18}&27.76 &\textbf{34.96}  & \textbf{76.41}\\
    \bottomrule
    \end{tabular}
        \caption{Comparison of preliminary document retrieval and answer results between IP-RAR and commonly used methods on the BioCDQA dataset. The best results are highlighted in bold.}
           \label{tab:bioCDQA} 
\end{table*}

\subsection{Performance of IP-RAR on Different Datasets}
For the BioASQ and MASH-QA datasets, we adhere to the evaluation criteria established by their respective datasets. 
\subsubsection{BioASQ dataset}

For the BioASQ dataset, Table~\ref{subtab:retrieval} demonstrates the superior document retrieval performance of IP-RAR. With higher precision and F1-score, IP-RAR effectively filters out irrelevant documents, while its leading performance in MAP and GMAP highlights its strength in ranking relevant documents. In contrast, although SELF-RAG achieves a higher recall, IP-RAR’s precision-focused trade-off results in better overall retrieval quality. Methods like bioinfo-0 and dmiip3 improve recall but perform worse in precision and ranking, often retrieving excessive irrelevant content.

\vspace{1em} 
It is noteworthy that both IP-RAR and SELF-RAG employ similar workflows: an ideal answer is first generated through their respective systems, followed by refinement using DeepSeek-V3 to extract the final response tailored to the specific question type. For these two systems, all evaluations are conducted on the extracted answers to ensure consistency and comparability. Other methods, such as dmiip3 and BioASQ Baseline, are evaluated based on their original outputs to maintain fairness and a consistent basis for comparison.
Table~\ref{subtab:answers} presents a comparison of exact answer results across IP-RAR and other methods. 
For Yes/No questions, IP-RAR achieves an F1 score and accuracy close to 96\%, second only to IISR-2. In Factoid questions, it ranks first with a Strict Accuracy of 68.36\% and an MRR of 67.34\%. For List questions, IP-RAR outperforms all systems with the highest Precision, Recall, and F1, demonstrating superior retrieval completeness.

\begin{table}[t]
  \centering
  
  \begin{subtable}[t]{\textwidth}
    \centering
    \begin{tabular}{@{}llllll@{}}
      \toprule
      \textbf{System} & \textbf{P (\%} & \textbf{R (\%)} & \textbf{F1 (\%)} & \textbf{MAP (\%)} & \textbf{GMAP (\%)} \\
      \midrule
      A\&Q4 & 10.27 & 58.16 & 17.46 & 44.04 & 2.15 \\
      dmiip3 & 11.33 & 61.27 & 19.12 & 44.62 & 2.40 \\
      bioinfo-0 & 21.18 & 60.47 & 31.37 & 45.90 & 2.67 \\
      SELF-RAG & 49.82 & \textbf{74.53} & 59.72 & 89.33 & 22.15 \\
      \textbf{IP-RAR} & \textbf{85.87} & 57.95 & \textbf{69.20} & \textbf{95.12} & \textbf{35.66} \\
      \bottomrule
    \end{tabular}
    \subcaption{}
    \label{subtab:retrieval}
  \end{subtable}

  \begin{subtable}[t]{\textwidth}
    \centering
    \begin{tabular}{@{}lccccccc@{}}
      \toprule
      \multirow{2}{*}{\textbf{System}} & \multicolumn{2}{c}{\textbf{Yes/No}} & \multicolumn{2}{c}{\textbf{Factoid}} & \multicolumn{3}{c}{\textbf{List}} \\
      \cmidrule(lr){2-3} \cmidrule(lr){4-5} \cmidrule(lr){6-8}
      & F1 & Acc. & Str. Acc. & MRR & Prec. & Rec. & F1 \\
      \midrule
      Baseline & 60.00 & 46.67 & 9.09 & 11.36 & 11.85 & 27.84 & 16.13 \\
      dmiip3 & 85.71 & 87.30 & 31.82 & 39.92 & 28.51 & 24.64 & 22.32 \\
      UR-gpt4 & 94.74 & 95.64 & 54.55 & 56.82 & 37.42 & 43.69 & 38.28 \\
      IISR-2 & \textbf{100.00} & \textbf{100.00} & 54.55 & 59.09 & 50.99 & 35.77 & 39.80 \\
      SELF-RAG & 85.14 & 82.55 & 30.61 & 28.31 & 51.10 & 22.38 & 29.19 \\
      \textbf{IP-RAR} & 95.91 & 95.34 & \textbf{68.36} & \textbf{67.34} & \textbf{88.78} & \textbf{54.53} & \textbf{63.04} \\
      \bottomrule
    \end{tabular}
    \subcaption{}
    \label{subtab:answers}
  \end{subtable}
    \caption{Comparison of document retrieval and exact answer results between IP-RAR and commonly used methods on the BioASQ dataset. (a) Document retrieval performance comparison. (b) Exact answer generation performance comparison. The best results are highlighted in bold.}
  \label{tab:bioasq_comparison}
\end{table}

\vspace{1em} 
In summary, IP-RAR outperforms other methods on the BioASQ dataset in both document retrieval and exact answer tasks, demonstrating superior precision, recall, ranking, and comprehensiveness.

\subsubsection{MASH-QA dataset}
 
 Table~\ref{tab:MASH} highlights IP-RAR’s superior performance on the MASH-QA dataset. For sentence retrieval, it achieves the highest F1 score (64.44\%) with a strong balance of Precision and Recall, outperforming SELF-RAG, which suffers from low Precision (27.46\%) despite its high Recall.
\vspace{1em} 

In answer prediction, IP-RAR’s Exact Match (47.29\%) more than doubles that of MultiCo, demonstrating superior accuracy. Notably, since both IP-RAR and SELF-RAG rely on LLMs to generate answers, the EM metric was evaluated using a GPT-4o-based \cite{hurst2024gpt} scoring system, which assigns a score of 1 only when the predicted answer and the gold standard convey the exact same meaning without any extraneous sentences. Any discrepancy in meaning or the inclusion of irrelevant sentences results in a score of 0. This strict scoring criterion highlights IP-RAR’s ability to produce concise and semantically accurate answers.
\vspace{1em} 

Compared to baseline models like BERT, RoBERTa, and XLNet, which exhibit low EM scores (below 10\%), IP-RAR excels in aligning retrieved sentences with precise answers. 
\begin{table}[t]
    \centering
    \begin{tabular}{lcccccc}
        \toprule
        \textbf{Model name} & \multicolumn{3}{c}{\textbf{Sentence}} & \textbf{Answer} \\
        \cmidrule(lr){2-4} \cmidrule(lr){5-5}
         & \textbf{P} & \textbf{R} & \textbf{F1} & \textbf{EM} \\
        \midrule
        TANDA & 56.48 & 16.42 & 25.44 & 8.95 \\
        BERT & 56.18 & 16.25 & 25.21 & 8.89 \\
        RoBERTa & 57.70 & 19.06 & 28.65 & 9.40 \\
        XLNet & 56.05 & 19.73 & 29.19 & 9.09 \\
        MultiCo & 58.16 & 55.90 & 57.00 & 22.05 \\
        SELF-RAG & 27.46 &\textbf{82.84}  &  41.26& 12.96 \\
        \textbf{IP-RAR(ours)} & \textbf{60.95} & 68.35 & \textbf{64.44} & \textbf{47.29}\\
        \bottomrule
    \end{tabular}
    \caption{Comparison of preliminary sentence retrieval results between IP-RAR and commonly used methods on the MASH-QA dataset. The best results are highlighted in bold.}
    \label{tab:MASH}
\end{table}

\subsection{Analysis of the Recall for Multi-Level and Multi-Granularity Retrieval Strategy}
Table~\ref{tab:recall} presents the recall (\%) results of the Multi-Level and Multi-Granularity Retrieval Strategy and its ablation variants on the BioCDQA dataset. From the multi-level perspective, combining abstract-level and full-text-level retrieval ensures a more comprehensive capture of relevant information, covering knowledge sources from concise to detailed. Multi-level retrieval improves recall by 8.76\% and 11.99\% for Question-based Retrieval and Virtual Answer-based Retrieval, respectively, demonstrating that single-level retrieval alone is insufficient to achieve full coverage. By integrating results across levels, multi-level retrieval significantly enhances performance. However, for Keyword-based Retrieval, multi-level integration shows limited improvement, as keyword-based methods struggle to retrieve meaningful information at the abstract level. This indicates that keywords are less effective in abstracts, but their strong performance at the full-text level compensates for this limitation.
\vspace{1em} 

From the multi-granularity perspective, the strategy combines question-based, virtual answer-based, and keyword-based retrieval to leverage different representations of relevance. The ablation results demonstrate that each granularity contributes effectively to the final outcome. Question-based retrieval achieves a recall of 40.04\%, while virtual answer-based retrieval further improves it to 44.06\%. Keyword-based retrieval, particularly with synonym expansion, significantly boosts recall to 66.10\%. This highlights that keyword-based retrieval complements the other granularities by capturing additional variations in linguistic expression. Importantly, each granularity contributes meaningfully, and their combined strengths play a crucial role in achieving the highest recall.
\vspace{1em} 

In summary, the integration of multi-level and multi-granularity strategies ensures the retrieval process captures relevant content comprehensively and effectively. The synergy between these dimensions is critical for achieving high recall, especially in large-scale datasets like BioCDQA, where maximizing coverage is essential for downstream applications.

\begin{table}[htbp]
\centering
\begin{tabular}{@{}lcccc>{\raggedright\arraybackslash}p{4cm}@{}}
\toprule
\multirow{2}{*}{Method} & \multicolumn{3}{c}{Hierarchical Retrieval} & \multicolumn{2}{c}{Multi-Granularity Retrieval} \\
\cmidrule(lr){2-4} \cmidrule(lr){5-6}
 & Abstract & Full-Text & Multi-Level & Value & Strategy Composition \\
\midrule
Question-based & 36.80 & 31.28 & 40.04 & 40.04 & \makecell{Only Question-based} \\
Virtual Answer-based & 34.79 & 27.63 & 39.62 & 44.06 & \makecell{+ Virtual Answer-based} \\
Keyword-based & 4.10 & 42.46 & 42.46 & 66.10 & \makecell{+ Keyword-based} \\
\bottomrule
\end{tabular}
\caption{\label{tab:recall}Recall (\%) for Multi-Level and Multi-Granularity Retrieval Strategy.}
\end{table}

\subsection{Impact of IP-RAR Components on Performance}

The ablation study presented in Table~\ref{tab:Ablation_IP-RAR} provides insights into the contributions of various components in the IP-RAR framework under the DeepSeek-V3. By isolating individual components, the analysis demonstrates how each affects the overall system's ability to generate accurate and contextually appropriate answers.
\vspace{1em} 

In the ablation study, \textit{w/o} Retrieval examines the system's performance when bypassing document retrieval entirely, relying solely on the DeepSeek-V3 to generate answers. In this configuration, the Precision, Recall, and F1 score metrics are not applicable, as no documents are retrieved. This setup leads to a significant performance drop, with a GPT-4 evaluation score of only 37.12\%. These results underscore the vital role of the retrieval process in grounding the model's answers with relevant context. Without retrieval, the system lacks access to supporting information, resulting in vague or less accurate answers.
\vspace{1em} 

For \textit{w/o} Integrated Reasoning-based Retrieval, the system bypasses the multi-level and multi-granularity retrieval strategy, directly retrieving a flat set of the top 50 text chunks using Contriever-MS MARCO and ranking the top 5 chunks during the progressive reasoning-based generation process. In this configuration, the Precision is 27.95\%, Recall is 18.04\%, and F1 score is 21.29\%. This results in a GPT-4 evaluation score of 50.18\%, reflecting the importance of the multi-level and multi-granularity retrieval strategy in refining the retrieval of highly relevant text chunks. While this alternative method retrieves some relevant documents, the overall recall remains too low to support comprehensive and accurate answer generation.
\vspace{1em} 

The \textit{w/o} Progressive Reasoning-based Generation ablation removes the Progressive Reasoning-based Generation module and generates answers directly from the content retrieved by the Integrated reasoning-based Retrieval module. In this setup, the Precision drops to 16.13\%, Recall increases to 32.81\%, and F1 score decreases to 21.62\%. Without the Progressive Reasoning-based Generation process, the system generates answers based on the top 5 chunks identified by the recall process, resulting in a GPT-4 evaluation score of 52.36\%. Experimental results show that retrieving too much irrelevant information either introduces large amounts of invalid knowledge that disrupts the generation process or causes the context to exceed the token limit of the LLM, leading to the removal of potentially crucial information. 
The complete IP-RAR framework, with all components integrated, achieves the highest performance, with a GPT-4 evaluation score of 76.41\%. Specifically, the Precision is 47.18\%, Recall is 27.76\%, and F1 score is 34.96\%. This result underscores the importance of each component in ensuring the system's ability to deliver precise and contextually grounded answers. Removing any component results in a significant performance degradation, affirming that retrieval, integrated reasoning-based retrieval, and progressive reasoning-based generation are all essential for achieving optimal results in biomedical question answering.

\begin{table}[t]
    \centering
    \begin{tabular}{lcccccc}
    \toprule
    \multirow{2}{*}{\textbf{System}} & \multicolumn{3}{c}{\textbf{Document Retrieval}} & \textbf{Answer} \\
    \cmidrule(lr){2-4} \cmidrule(lr){5-5}
     & \textbf{Prec. (\%)} & \textbf{Rec. (\%)} & \textbf{F1 (\%)}  & \textbf{GPT-4 Eval (\%)} \\
    \midrule
    IP-RAR(ours)  & 47.18 & 27.76 & 34.96 & 76.41 \\
    \hline
    \textit{w/o} Progressive Reasoning-based Generation & 16.13 & 32.81 & 21.62  & 52.36 \\
    \textit{w/o} Integrated Reasoning-based Retrieval & 27.95 & 18.04 & 21.29  & 50.18 \\
    \textit{w/o} Retrieval & / & / & /  & 37.12 \\
    \bottomrule
    \end{tabular}
        \caption{Ablation study of the IP-RAR framework}
         \label{tab:Ablation_IP-RAR}
\end{table}


\subsection{Formulating Scientific Questions and Planning Research}

Formulating scientific questions and designing research plans are critical steps in driving innovation and breakthroughs in biomedical research. A well-defined and challenging scientific question not only determines the direction of the study but also influences data collection, selection of experimental methods, and the interpretability of research outcomes. A well-structured research plan, in turn, optimizes resource allocation and enhances both feasibility and impact. By efficiently leveraging existing articles to identify research gaps and propose novel scientific hypotheses, researchers can significantly improve research efficiency. These hypotheses not only provide a solid theoretical foundation for subsequent experiments or clinical studies but also accelerate the understanding of disease mechanisms and the development of novel therapeutic strategies. Systematic analysis and integration of articles enable researchers to uncover unknown aspects of disease biology or limitations in current therapeutic approaches, thereby improving the rationality of hypotheses and the relevance of experimental designs. This process is crucial for drug development and the optimization of precision medicine strategies, ultimately expediting scientific discovery \cite{rajpal2014mining, mohs2017drug}.

\vspace{1em}

Our framework facilitates the efficient extraction of key insights from existing article, generating scientific hypotheses and designing research plans. Figure~\ref{fig:scientific_questions} illustrates the systematic process of retrieval, integration, and reasoning, which ultimately generates evidence-based scientific questions to guide subsequent experimental designs. For instance, miR-375 may play distinct roles in the progression of different colorectal cancer subtypes, such as adenocarcinoma, squamous cell carcinoma, and small cell lung cancer. Its target gene, ITPKB, may be directly regulated by miR-375 and play a crucial role in cancer development. Based on this hypothesis, researchers can conduct targeted in vitro experiments, such as luciferase reporter assays and Western blot analysis, to validate the regulation of ITPKB by miR-375. Additionally, in vivo studies using mouse models can effectively evaluate the impact of miR-375 on tumor growth and metastasis across different cancer subtypes. Further, large-scale high-throughput analyses of miR-375 expression patterns and its target genes in colorectal cancer patient cohorts, combined with correlations to clinical outcomes, can provide actionable research insights. This structured approach not only guides experimental design but also optimizes research workflows, enhancing both rigor and efficiency. By leveraging this literature-driven research methodology, researchers can accelerate scientific discoveries and improve the planning and execution of biomedical studies.

\subsection{Drug Interaction Research in Clinical Decision Support}

Research on drug interactions plays a crucial role in clinical decision support, ensuring treatment safety and efficacy, particularly in the context of polypharmacy and personalized medicine \cite{maher2014clinical}. Many patients require multiple medications to manage comorbidities, making a deep understanding of drug interactions essential for clinicians to identify potential risks, such as enhanced side effects, altered drug metabolism, or reduced therapeutic effectiveness. Leveraging our approach, clinicians can more efficiently uncover synergistic effects that enhance therapeutic outcomes or identify antagonistic interactions that compromise efficacy \cite{neuvonen2006drug}. This process not only aids in preventing adverse reactions but also supports the development of safer and more effective treatment plans, especially for elderly patients and those with multiple comorbid conditions.
\vspace{1em} 

\begin{figure}[]
  \centering
  \begin{subfigure}[b]{0.8\textwidth}
    \centering
    \includegraphics[width=\linewidth]{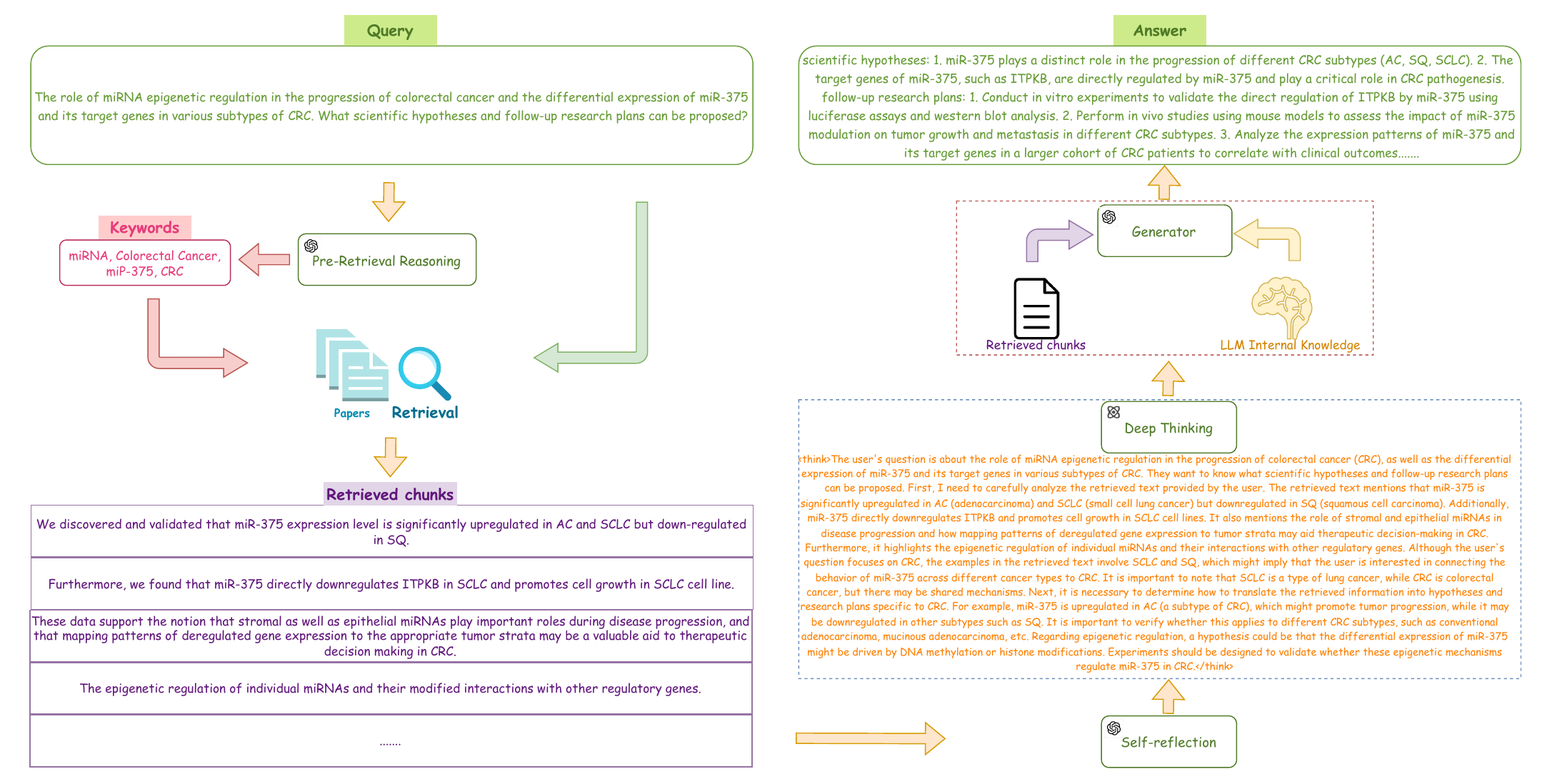}
    \caption{}
    \label{fig:scientific_questions}
  \end{subfigure}
  \begin{subfigure}[b]{0.8\textwidth}
    \centering
        \includegraphics[width=\linewidth]{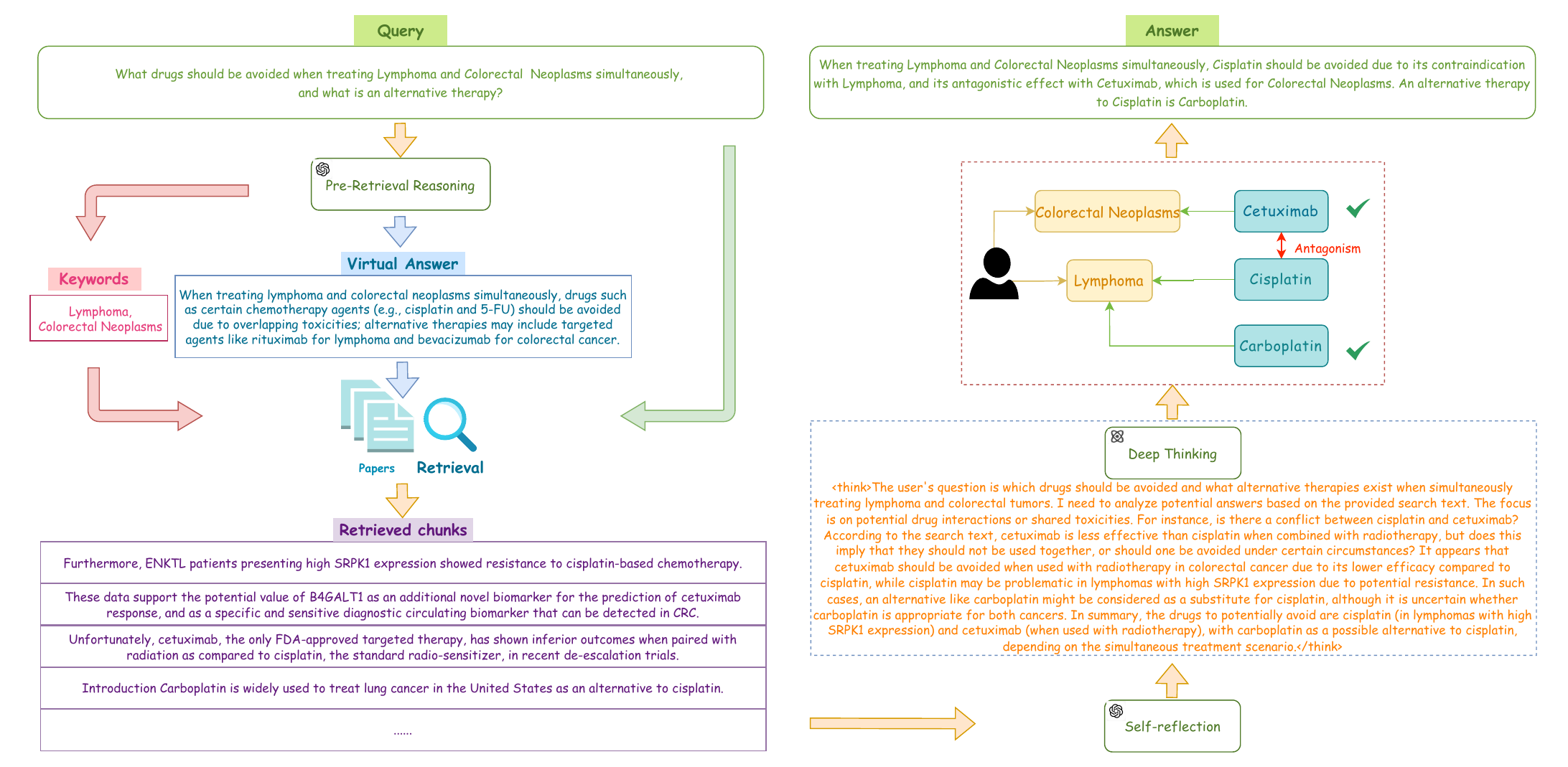}
    \caption{}
    \label{fig:drug_interaction}
  \end{subfigure}
  \caption{Examples of Applications in Biomedical Research and Clinical Decision Support. (a) An Example of Formulating Scientific Questions and Planning Research. (b) An Example of Drug Interaction Research for Clinical Decision Support.}
  \label{fig:research_examples}
\end{figure}

For patients with coexisting lymphoma and colorectal cancer, robust clinical decision support is essential for optimizing treatment plans and reducing the risks associated with drug interactions. Figure~\ref{fig:drug_interaction} illustrates the system’s retrieval and reasoning process, ultimately generating evidence-based answers.
IP-RAR first performs pre-retrieval reasoning using a large language model, extracting keywords from the query and generating an initial response. It then retrieves relevant medical articles from databases to extract supporting evidence, including the facts that Cisplatin may induce resistance in lymphoma and that it has an antagonistic interaction with Cetuximab. The system then applies self-reflection to determine which text chunks are relevant to the question, followed by deep thinking to infer drug interactions. Finally, it recommends avoiding Cisplatin and suggests Carboplatin as an alternative therapy. This evidence-based reasoning approach assists clinicians in developing safer and more effective treatment plans, thereby enhancing patient safety and improving outcomes for individuals with complex, multi-disease conditions.

\section{Conclusion}
To advance biomedical knowledge extraction and application, this paper proposes a comprehensive framework that integrates knowledge graphs with LLMs. By constructing the BioStrataKG from large-scale biomedical articles, the framework systematically uncovers multi-layered relationships among biomedical entities, such as genes, proteins, diseases, and drugs, as well as interactions between research papers in terms of methodologies, datasets, research directions, and citation relationships. To support cross-document reasoning and biomedical knowledge mining, we develop the BioCDQA dataset based on BioStrataKG, addressing the limitations of existing biomedical QA datasets in handling cross-document and high-level reasoning tasks. Furthermore, we introduce the IP-RAR framework, which combines Integrated Reasoning-based Retrieval with Progressive Reasoning-based Generation, enabling LLMs to efficiently retrieve, synthesize, and utilize multi-source evidence.
\vspace{1em} 

Experimental results demonstrate that IP-RAR significantly outperforms previous approaches in both retrieval efficiency and answer accuracy. In clinical applications, the framework assists physicians in rapidly identifying and synthesizing critical information from vast biomedical articles, facilitating the development of more precise personalized treatment plans. In research settings, it enables systematic analysis of cutting-edge advancements and identification of potential research gaps, accelerating research strategy formulation and decision-making. IP-RAR holds potential in various biomedical domains, including drug synergy/antagonism analysis, drug repurposing, and precision medicine, providing essential technical support for advancing biomedical research and clinical practice.
\vspace{1em} 

Despite its promising results, IP-RAR still faces challenges in handling highly complex multimodal data, dynamically evolving scientific knowledge, and human-AI interaction. Future work will explore multimodal information integration and agent-based interactions to further enhance the robustness and adaptability of the framework, ensuring more comprehensive support for biomedical knowledge mining and clinical applications. By continuously improving the dynamic maintenance of knowledge graphs and refining reasoning mechanisms, the proposed framework is expected to establish a more efficient and precise paradigm for biomedical knowledge discovery, fostering rapid advancements in scientific research and clinical decision-making.

\section{Experimental Section}
\subsection{Construction of the BioStrataKG}

Here, we detail the construction process of the document-entity dual-layer representation fusion architecture—BioStrataKG, as illustrated in Figure \ref{fig:dataset_construction}. Initially, a dual knowledge extraction strategy is employed: the GPT-4o mini is utilized for fine-grained semantic parsing to extract entity-relation triplets from the literature, constructing an entity-level knowledge graph; simultaneously, the GPT-4o mini is used for macro semantic understanding to extract document-level meta-knowledge such as research methods and directions, building a document knowledge graph. By integrating these two knowledge graphs, we establish a cross-literature knowledge association network, effectively supporting hierarchical reasoning and cross-document correlation analysis in biomedical knowledge.
\subsubsection{Data Collection and Processing}
\threesubsection{Paper Collection}\\
The papers included in our dataset are all sourced from the open-access database PubMed, which allows anyone to download and access biomedical and life sciences article via PMID. We have downloaded over 8,000,000 papers, extracting their titles, abstracts, publication years, keywords, and citation relationships.

\threesubsection{Paper selection}\\
We filtered papers based on keywords related to lung cancer, breast cancer, and colorectal cancer, focusing on studies involving single-cell analysis, pharmacology, and clinical trials. We further narrowed our selection to papers published within the last decade and those with more than 10 citations. In total, we selected 15,585 papers to build the knowledge graph for generating the bioCDQA dataset. 
Some of the basic statistics of the papers involved in our dataset are detailed in Supporting Information Fig. S3.

\threesubsection{PDF to Markdown Conversion}\\
Since the papers downloaded from the database are in PDF format, we convert them to Markdown format \cite{marker_repo,DBLP:journals/pr/WangHZSH24}.

\subsubsection{Knowledge Graph Construction}
We leverage the GPT-4o mini to extract biomedical-related triplet information (e.g., genes, proteins, diseases, drugs, etc.) from research papers to construct an entity-level knowledge graph. Simultaneously, we structurally summarize the research methods, datasets used, and research domains of the papers to build a document-level knowledge graph. On this basis, each paper serves as a node within the document-level knowledge graph. By identifying recurring biomedical terms, related datasets, research methods across multiple papers, and analyzing citation and reference relationships, we establish inter-document linkages and achieve comprehensive knowledge integration. This cross-document information connection not only facilitates the construction of high-quality cross-document question-answering datasets but also provides richer contexts and deeper opportunities for knowledge discovery in biomedical knowledge mining tasks. Additionally, we have open-sourced this knowledge graph, which will soon be made publicly available.
Beyond supporting our dataset, the knowledge graph can serve as a valuable resource for researchers, aiding in knowledge discovery, identifying trends, and exploring relationships between biomedical entities and research methods. The knowledge graph contains 94,962 nodes and 290,403 relationships. Detailed descriptions of the nodes and relationships are provided Figure \ref{fig:rag_KG}\\

\threesubsection{Entity-Level Knowledge Graph}\\

\noindent\textbf{Triplet Extraction Based on LLMs}. In the KG, a triplet is the fundamental unit of information used to represent entities and their relationships. Specifically, a triplet is defined as \((e_1, r, e_2)\), where \(e_1\) and \(e_2\) are entities, representing nodes in the knowledge graph, and \(r\) is a relation, representing the connection between the entities. 
We utilize the GPT-4o mini to extract triplets from the abstracts of papers. The entity types (\(ET\)) we extract are Gene, Protein, Drug, and Disease. 
The relation types (\(RT\)) we define and Triplet Extraction Prompt are detailed in Supporting Information Fig. S4. 

\noindent\textbf{Entity Normalization in Triplet Extraction}.
Entity normalization is essential for ensuring consistent and standardized representation of extracted entities in knowledge graph construction. To address this, we adopt a two-stage approach that combines domain-specific databases with advanced language models. First, we reference the MeSH database \cite{nlm_mesh_2023} for diseases and drugs, and the UniProt database \cite{uniprot2023uniprot} for genes and proteins. Using the all-MiniLM-L6-v2 \cite{wang2020minilm} embedding model, we retrieve the top 5 candidate terms from these databases by generating dense vector representations of the extracted entities, ensuring computational efficiency and high recall. Next, GPT-4o mini evaluates the semantic and contextual alignment between the extracted entity and the retrieved candidates to identify the most appropriate standardized term. This approach resolves ambiguities, such as homonyms or abbreviations, and ensures that entities are contextually accurate and semantically consistent.

\noindent\textbf{Constructing Entity-Level Knowledge Graphs}. Once the standardized triplets \((e_1, r, e_2)\) are obtained, the next step is to construct the entity-level knowledge graph, where entities \(e_1\) and \(e_2\) are represented as nodes and the relation \(r\) forms the directed edge between them. The process begins with node creation, where each unique entity from the standardized triplets is instantiated as a node in the graph and categorized based on its type (e.g., Gene, Protein, Drug, Disease) to ensure consistency and facilitate downstream analysis. Relations are then represented as directed edges, labeled with their relation types, connecting the corresponding nodes and formalizing the interactions between entities. The graph supports multiple labeled edges between the same nodes, capturing diverse relationships such as a gene being associated with multiple diseases, and manages complexity by representing each distinct relationship as a separate edge. During the construction process, rigorous checks are applied to detect and handle duplicate nodes and edges, ensuring that new relationships are accurately linked to existing nodes, thereby preventing redundant entries. The completed KG is stored in the Neo4j graph database, where Neo4j’s Cypher query language enables efficient and intuitive exploration of the graph. Users can retrieve complex relationship networks, such as querying all drugs related to a specific disease or identifying potential relationship chains between entities, facilitating flexible and powerful knowledge discovery.\\

\threesubsection{Document-Level Knowledge Graph}\\

\noindent\textbf{Paper Information Extraction Based on LLMs}
We utilize the GPT-4o mini to extract the fundamental research methods of each paper, the datasets used, and the respective research domains. These extracted elements, along with the paper titles, are considered as nodes. Each node and its relationship constitute a document-level triplet \((e_1, r, e_2)\), facilitating subsequent knowledge graph construction. The prompt for extraction of paper information is presented in Supporting Information Fig. S5.

\noindent\textbf{Normalization.}
To achieve standardization and consistency of entity names in document-level knowledge graphs, we propose a standardization workflow based on vector matching and LLM-assisted decision-making. The workflow processes each entity name in the triples sequentially, ensuring that all potential duplicate entities are incorporated into subsequent matching and standardization steps. For each entity name, the existing vector database (e.g., ChromaDB) is traversed to calculate the cosine similarity with other entities. If a record with a similarity score greater than 0.5 is found, the two entity names are considered to refer to the same concept (e.g., a method or dataset), and the current entity name is merged with the existing record. Conversely, if no matching entity name is found, the current entity name is inserted into the vector database for use in future standardization steps.
Building on the initial steps, GPT-4o mini further optimizes merged entity names by identifying the most semantically relevant and contextually accurate choices, ensuring standardization, precision, and robustness in the final output.

\noindent\textbf{Constructing Document-Level Knowledge Graphs from Triplets.}  
After normalizing document-level triplets \((e_1, r, e_2)\), the document-level knowledge graph is constructed in a manner similar to the entity-level KG. Each unique entity, including papers, research methods, datasets, and research domains, is instantiated as a node in the graph. Directed edges are created between these nodes to represent the relationships, ensuring an efficient and comprehensive structure for knowledge discovery.

\subsection{Construction of the Biomedical Cross-Document Question Answering Dataset}
Based on the BioStrataKG, we proposed a new biomedical question-answering dataset, BioCDQA, to support cross-document reasoning and biomedical knowledge mining. The construction of the dataset is illustrated in Figure \ref{fig:rag_dataset}. 
The dataset construction process integrates both unstructured data from raw text and structured data from knowledge graphs for biomedical research. By extracting information from unstructured text, we capture diverse, contextually rich data, while structured knowledge graphs provide precise, well-organized relationships between entities such as genes, diseases, drugs, and proteins. These complementary sources ensure a robust dataset capable of supporting a variety of downstream tasks. The statistics of various question categories are summarized in Figure \ref{fig:Statistical Analysis} (b). BioCDQA integrates knowledge graphs with LLMs and combines automated generation with manual selection to ensure the quality and diversity of question-answer pairs, thereby better supporting complex knowledge reasoning tasks.\\ 

\subsubsection{Dataset Generation from Unstructured Data}

We employ the BioStrataKG to establish relationships among multiple research papers. Randomly selecting 1 to 5 interconnected articles, we input the full text of the chosen papers into GPT-4o mini. This system then formulates pertinent questions from the following perspectives and generates responses based on both the questions and the text.

\subsubsection{Dataset Generation from Structured Data}

We generate various Cypher query statements based on BioStrataKG, which features two distinct granularity-based knowledge graphs (entity-level and document-level), to extract relevant subgraphs and relationship chains. The Entity-Level Knowledge Graph provides a comprehensive framework for exploring complex biomedical interactions, such as those between diseases, genes, drugs, and proteins. This supports key research areas, including drug repurposing, gene-disease prediction, and treatment optimization. Meanwhile, the Document-Level Knowledge Graph is designed to enhance research efficiency by offering tailored recommendations aligned with specific research directions. These queries retrieve nodes and relationships pertinent to specific research topics, enabling the extraction of focused subgraphs and relationship chains. By leveraging these, GPT-4o mini formulates targeted questions and generates corresponding answers, improving question precision and dataset quality.

\subsubsection{Dataset Inspection with LLM}

Leveraging both unstructured and structured data, we utilize the capabilities of GPT-4o mini to generate a substantial corpus of over 5,000 question-answer pairs. To verify the reliability of the generated answers, we then employ GPT-4o. Given the original source articles, we design prompts for GPT-4o to identify the specific passage within the article that supports each QA pair. Dataset Inspection Prompt are detailed in Supporting Information Fig. S6. \\

\subsubsection{Manual Inspection and Selection}

We curated and manually selected 1,183 QA pairs to create bioCDQA, a dataset designed for biomedical research and applications. Retrieval involves navigating a large corpus of 68,428 research papers, reflecting the complexity of the knowledge space. Each QA pair links questions to precise biomedical concepts and relationships, facilitating deep knowledge mining across multiple documents and providing a high-quality, contextually relevant resource for advanced reasoning and question answering.

\subsection{Integrated and Progressive Retrieval-Augmented Reasoning Framework}
\label{sec:IP-RAR}
Here, the IP-RAR framework is introduced for deep knowledge mining and question answering in biomedical articles. Figure \ref{fig:frameworks} illustrates the framework. 
\subsubsection{Integrated Reasoning-based Retrieval}
\threesubsection{Pre-Retrieval Reasoning}\\  
The first step employs the DeepSeek-V3 \cite{liu2024deepseek} to perform pre-retrieval reasoning by extracting key terms to identify relevant biomedical entities and generating synonyms to enhance matching flexibility. This approach overcomes the limitations of existing methods that struggle to precisely locate answers in complex queries, enabling more accurate targeting of specific biomedical entities. Furthermore, based on the pre-retrieval reasoning process, the system generates a virtual answer as a hypothesis to refine retrieval precision, improving the identification of relevant information. This improves the system’s ability to link related paragraphs and retrieve contextually consistent information, with the prompt for pre-retrieval reasoning detailed in Supporting Information Fig. S7.

\threesubsection{Multi-Level and Multi-Granularity Retrieval Strategy}\\ 
To enhance recall and maximize the retrieval of relevant knowledge from millions of text chunks, we employ a retrieval strategy that integrates multi-level and multi-granularity approaches. The multi-level aspect combines abstract-based and full-text-based retrieval to capture relevant content across varying depths of detail. The multi-granularity aspect involves question-based, keyword-based, and virtual answer-based retrieval, ensuring a comprehensive exploration of the corpus. 
Using the Contriever-MS MARCO model, we first retrieve the top 10 abstracts based on the similarity between the question and text chunks, followed by another top 10 abstracts based on the similarity between the virtual answer and text chunks. At the full-text level, we retrieve the top 10 chunks for the question and an additional top 10 for the virtual answer using the same model. Additionally, we apply keyword-based matching techniques, expanding retrieval by including synonyms to increase flexibility. 
This multi-level and multi-granularity retrieval strategy effectively captures relevant content from different perspectives and depths, significantly improving the alignment of retrieved knowledge with the question's intent from a vast corpus.

\threesubsection{Aggregator}\\ 
The aggregator framework employs a weighted normalization function to rank retrieved chunks effectively. This function integrates three key factors: similarity score, method diversity, and intra-document repetition, ensuring contextually relevant and diverse rankings. Each factor is weighted according to its relevance, producing a normalized score \( S_i \) for each chunk, allowing for comparison across all retrieved chunks. The score for each chunk \( i \) is calculated as:

\begin{equation}
    S_i = w_S \times \frac{S_{\text{sim},i}}{S_{\text{sim},max}} + w_M \times \frac{M_{i}}{M_{max}} + w_R \times \frac{R_{i}}{R_{max}} 
    \label{eq:normalized_score}
\end{equation}

where \( S_{\text{sim},i} \) represents the similarity score of chunk \( i \), \( M_i \) indicates the number of distinct retrieval methods that identified chunk \( i \), and \( R_i \) reflects the count of retrieved chunks from the same document as chunk \( i \). The terms \( S_{\text{sim},\text{max}} \), \( M_{\text{max}} \), and \( R_{\text{max}} \) are the maximum values of these metrics across all chunks, used for normalization. The weights \( w_S \), \( w_M \), and \( w_R \) control the importance of each factor, where \( w_S \) emphasizes the relevance of similarity scores, \( w_M \) highlights chunks retrieved by multiple methods, and \( w_R \) prioritizes chunks with higher intra-document retrieval coverage. The resulting normalized score \( S_i \) provides a robust ranking metric that integrates query relevance, retrieval diversity, and intra-document coverage, providing a comprehensive and balanced ranking metric. In our experiments, these weights are set to \( w_S = 5 \), \( w_M = 3 \), and \( w_R = 1 \).

\subsubsection{Progressive Reasoning-based Generation}
The Progressive Reasoning-based Generation process in our proposed IP-RAR framework is designed to ensure that only the most relevant evidence is employed in generating the final response. This process unfolds in four distinct phases:

\threesubsection{Relevance Check}\\  
In this phase, all candidate text chunks are ranked in descending order based on their retrieval scores. The ranked chunks are then sequentially presented to the DeepSeek-V3, which evaluates whether each chunk sufficiently answers the posed question. This process continues until 5 relevant chunks are identified. If fewer than 5 relevant chunks are available, only those identified as relevant are used; if no relevant chunks are identified, the top 5 highest-scoring chunks are selected regardless of their assessed relevance.

\threesubsection{Answer Construction}\\
 After selecting the candidate text chunks, a simple prompt is constructed to allow the DeepSeek-V3 to generate an initial response to the question. The prompt incorporates the context of the question and the selected chunks to facilitate a coherent and informed answer.

\threesubsection{Self-Reflective Evaluation}\\
Following the generation of the response, a self-reflection evaluation is performed to systematically assess the degree to which each text chunk supports the generated response. The DeepSeek-V3 is prompted to evaluate the relevance of each chunk with respect to the question and the proposed answer, assigning a support score to each. Scores are given on a scale from 0 to 100, where 100 represents maximum relevance, and 0 indicates that the chunk is either entirely irrelevant or contradicts the proposed answer. Intermediate scores reflect varying levels of support based on the chunk's contribution to the response.

\threesubsection{Deep Thinking}\\
Based on the support scores assigned during the self-reflective evaluation, the system prioritizes the text chunks corresponding to the most relevant answers and utilizes DeepSeek-R1 to perform deep-thinking-based reasoning for the final answer generation. The DeepSeek-R1 is prompted to integrate these highly relevant text chunks, producing a cohesive and precise final response to ensure both contextual consistency and comprehensive accuracy.

\bibliographystyle{unsrt}
\bibliography{sn-bibliography}

\end{document}